\documentclass[10pt,twocolumn,letterpaper,fleqn]{article}
\usepackage{array,multirow}
\usepackage{cvpr}
\usepackage{times}
\usepackage{epsfig}
\usepackage{graphicx}
\usepackage{amsmath}
\usepackage{amssymb}
\usepackage{enumitem}
\usepackage{multirow}
\usepackage{graphicx}
\usepackage{caption}

\cvprfinalcopy % *** Uncomment this line for the final submission

\def\cvprPaperID{16} % *** Enter the CVPR Paper ID here

\begin{document}

%%%%%%%%% TITLE
\title{In Search of Life: Learning from Synthetic Data to Detect Vital Signs in Videos}

\author{Florin Condrea\\
Arnia Software, Romania\\
{\tt\small florin.condrea@arnia.ro}
\and
Victor-Andrei Ivan\\
Arnia Software, Romania\\
{\tt\small victor.ivan@arnia.ro}
\and
Marius Leordeanu\\
Arnia Software, Romania\\
University “Politehnica” of Bucharest\\
{\tt\small leordeanu@gmail.com}
}

\maketitle
%\thispagestyle{empty}

%%%%%%%%% ABSTRACT
\begin{abstract}
   Automatically detecting vital signs in videos, such as the
   estimation of heart and respiration rates, is a challenging research problem in computer vision with important applications in the medical field. One of the key difficulties in tackling this task is the lack of sufficient supervised training data, which severely limits the use of powerful deep neural networks. In this paper we address this limitation through a novel deep learning approach, in which a recurrent deep neural network is trained to detect vital signs in the infrared thermal domain from purely synthetic data. What is most surprising is that our novel method for synthetic training data generation is general, relatively simple and uses almost no prior medical domain knowledge. Moreover, our system, which is trained in a purely automatic manner and needs no human annotation, also learns to predict the respiration or heart intensity signal for each moment in time and to detect the region of interest that is most relevant for the given task, e.g. the nose area in the case of respiration. We test the effectiveness of our proposed system on the recent LCAS dataset and obtain state-of-the-art results.
\end{abstract}

%%%%%%%%% BODY TEXT
\section{Introduction}

Vital signs monitoring is an important part of the medical field, at the intersection between medicine and the fast development of technology. It is no longer a topic that is only available in hospitals, as significant technical improvements in wearable devices make it suitable for every day use at home. Moreover, advancements in cameras and other devices, in combination with powerful vision and machine learning algorithms, prove that emerging smart medical technologies are able to provide measurements that meet or even surpass the traditional medical gold standards [1, 2].

Part of the wide variety of camera sensors, thermal cameras, which sense the skin temperature distribution, which is correlated with various other body signals, could also constitute a good source for estimating breathing patterns [3,4], pulse [5] and even stress levels [6]. A specific vital sign rate is measured as the number of cycles per unit of time (usually per minute). Normal vital sign rates vary according to multiple factors, such as age, psychical fitness and health issues, leading to wide ranges of normal vital signal rates. For example, in the case of adults, normal respiratory rates lie within 12 and 20 breaths per minute, and normal heart rates between 60 to 100 beats per minute. Besides the normal sign pattern, abnormal patterns may also occur [7], as temporary cessation (Apnea), abnormally low rate (Bradypnea) or abnormally high rate (Tachypnea). All these factors of variation make the task difficult, while powerful deep learning models, which could address such challenges, cannot be easily used due to lack of supervised training data.

Vital signals can be detected using multiple cues. A standard signal source for estimating breathing rate, in the thermal domain, is the variation in heat around the nose due to inhaling cold air and exhaling hot air [8] - but for that approach we would need to know where the nose is in the image.
In the case of heart rate signal, information about the signal of interest can be detected by slight variation in face color [9] or superficial blood vessels [33], but there are many unrelated factors (e.g. illumination changes and other noises) that can affect these.

In our work, we propose to address all these limitations via a deep learning approach, with a recursive neural network, termed VSignNet, which learns, from synthetic data alone and without any human supervision, to predict both vital sign intensity and the corresponding regions of interest in thermal videos. Our method, to the best of our knowledge, is the first of its kind on this task, and achieves top results on the recent LCAS dataset [10], which is one of the very few ones available for this problem. VSignNet is applied directly on the input of thermal frames and predicts for each frame, along two output pathways, the value of the signal of interest and the region of interest that is likely to be the most important source of the signal e.g. the nose area for respiration. 

\begin{figure*}[!t]
  \includegraphics[width=\linewidth,height=6cm]{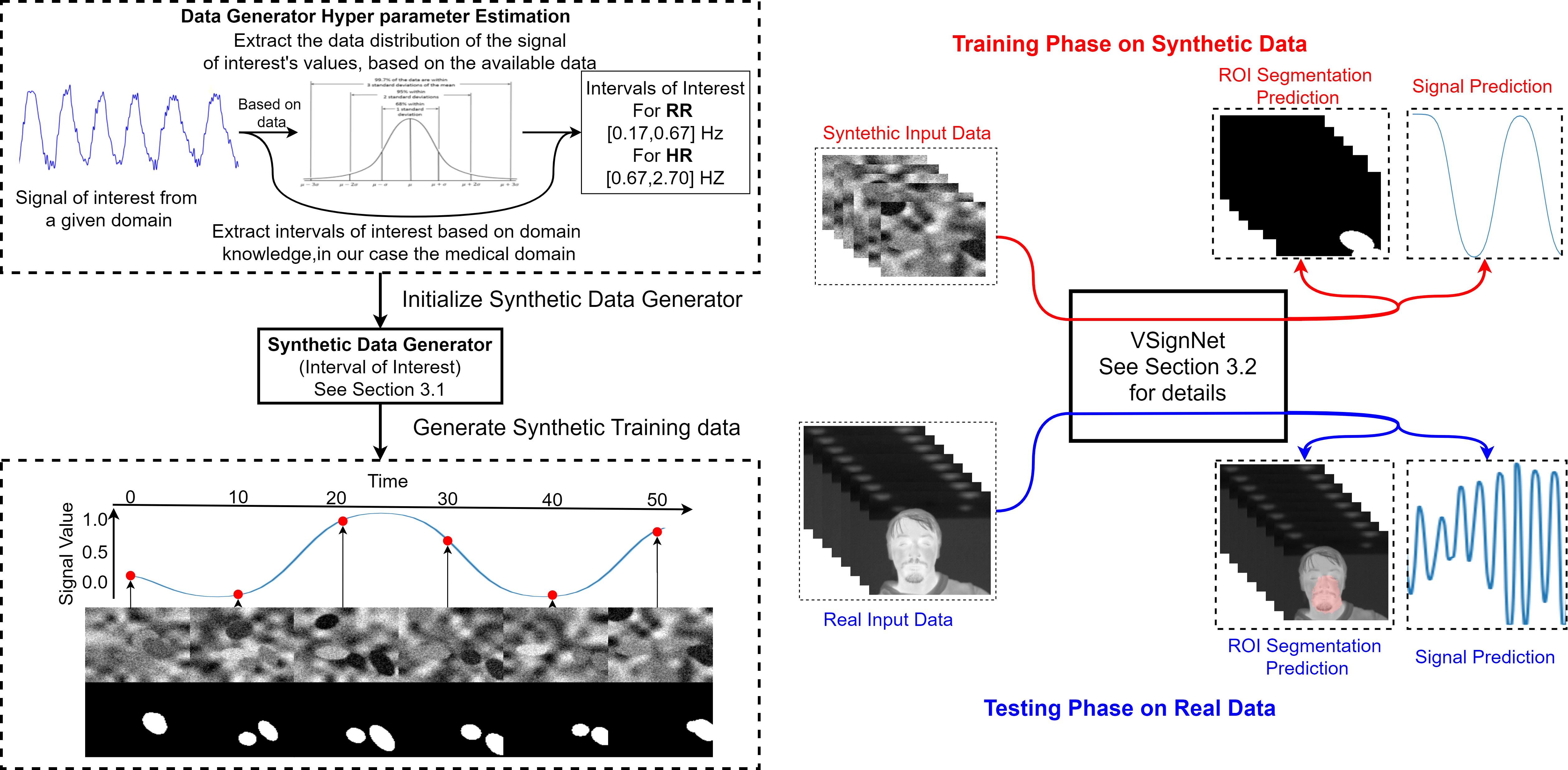}
  \caption{The four stages of our approach: the first stage consists of determining the frequency range of interest, either by using dataset statistics or domain knowledge. The second stage is the generation of synthetic training data samples. The third stage is the training of the model. At the final, inference stage, an additional post-processing step detects the peaks of the signal, on which we base our output average signal rate.}
  \label{fig:sample1}
\end{figure*}

\noindent \textbf{The main contributions introduced are:}
   \begin{enumerate}
        \item A novel deep learning approach trained without human supervision on synthetic data for detecting heart and respiratory signs in thermal videos with state-of-the-art performance on the LCAS dataset [10].
        \item Our deep-synthetic model is a able to estimate not only the vital sign rate, but is also learns, without any human supervision, to detect the intensity of the signal in every moment of time as well as the region of interest in the image corresponding to the signal.
        \item A general method for synthetic training data generation which uses minimal medical information and anatomical cues and no strong prior knowledge of the target signal frequency. 
   \end{enumerate}

%------------------------------------------------------------------------
\section{Scientific context and a baseline}

Important work has been developed for vital sign measurement in the RGB domain.In [28,29] authors present 
remote physiological measurement algorithms using signal analysis.In [30] authors introduces an 
algorithm using Independent Component Analysis to extract a signal of interest for heart rate, heart rate 
variability and breathing rate estimation. In [31] a Color Distortion Filter is introduced, which showed improved results when 
used as a pre-processing step with existing remote Photoplethysmography methods.In [32] 
introduce a novel convolutional attention network which recovers 
blood volume pulse and respiration signals from video, applicable on both RGB and Infrared.

The current literature for vital sign detection in thermal videos 
seems to revolve around a common experimental paradigm [10-12,14].
Usually, in the experimental setup, a small 
number of subjects is filmed with a thermal sensor. For example, in recent work [10], 
5 videos are recorded of 5 subjects are sitting in front of
the thermal camera. The environment in 
which the experiment is performed is mostly constrained, 
being indoors and lacking variability. Of course that such a small sample size and relatively limited 
experimental setup make it difficult to train powerful deep neural networks
for vital sign monitoring. That justifies  our approach of designing a method to synthetically generate  training data, which would enable the training of large deep networks.
The most common approach in the literature 
defines a pipeline with three steps: 

\textbf{The first step:} is the
detection of a region of interest (ROI), from which the signal of interest is extracted at each 
frame. For example, the approach in [12] 
detects the ROI (nostril region) in the first frame and tracks the corresponding ROI in the next frames
using the Median Flow algorithm [13]. They provide three possible ways for detecting the nostril
region in the first video frame: by manual initialization, by applying the method from [14] 
using human anatomy 
queues in the thermal image, or by cross-correlation between the thermal image and a database of nostril
ROI's. Alternatively, the method in [10] first segments the face in order to build a box around it, and then uses a pretrained
landmark detector to find the position of the ROI (nostril region). 

\begin{figure*}[!t]
  \includegraphics[width=\linewidth,height=4cm]{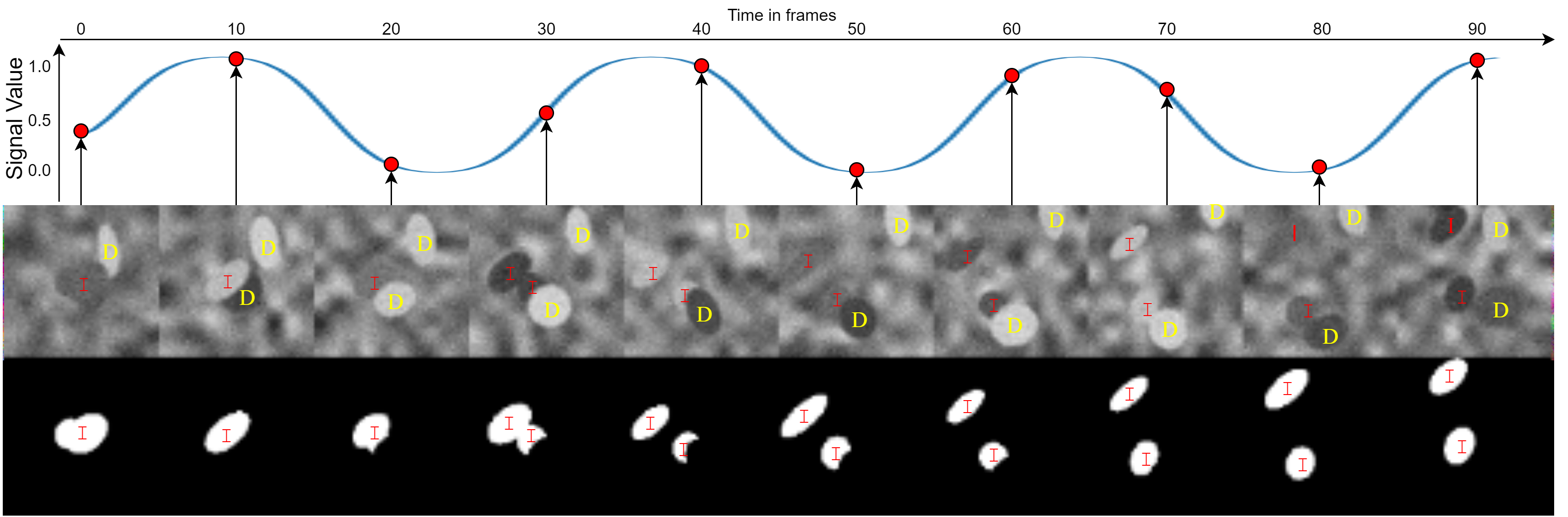}
  \captionsetup{justification=centering}
  \caption{
    Synthetically generated training data: target signal of interest $S_I$ (top row); 
    corresponding input frames (middle row) and ground truth ROI segmentations (bottom row),
    sampled every 10 frames. Input frames are created by combining objects of
    interest "I" (with intensity varying w.r.t $S_I$), distractor objects "D" (with intensity varying w.r.t their own signal $S_D^k$)
    and varying background $\beta_{BG}$. Finally, the synthetic frame is smoothed with a Gaussian ($GB_{\sigma}$) and 
    Salt-and-Pepper noise ($\beta_{SP}$) is added.
  }
  \label{fig:boat1}
\end{figure*}

\textbf{The next two steps:} are signal extraction and frequency computation. For signal extraction, 
recent methods [10,14] compute the mean pixel value inside the ROI, after the first ROI detection step.
In [11] pixel-wise signals are extracted to find the final
breathing time series, whereas in [12] authors consider the ROI pixels as voxels and compute the volume of the resulting shape at each frame. For frequency computation, the approaches in [11, 12] 
perform spectral analysis in the Fourier domain and, in [11], the authors 
detect peaks in the time series to find inhalation onsets. 

\textbf{Creating a strong baseline:}
In order to study the limitations of the ROI detection approach for breathing rate computation, 
we implemented a baseline method and tested it on the LCAS dataset, provided in [10]. 
For ROI detection we used RetinaFace [15],
a publicly available state-of-the-art face detector, trained on the WIDER FACE dataset[16]. RetinaFace outputs a bounding box and 5 facial landmarks for each face in the image. Out of the 5 landmarks, two are for the left and right eye, two for the corners of the mouth and one for the tip of the nose. In order to extract the nostril region, we select a box 8 times smaller than the face bounding box around the nose tip landmark (number determined experimentally). In this way the variance of the extracted signal does not change significantly when the person moves closer or farther away from the camera. We noticed that ROI detection can introduce noise into the extracted signal due to 
inaccurate detection, fluctuating in size from frame to frame. The accuracy of methods based on ROI (nostril) detection
is also sensitive to different cases that may appear in real life. For example, switching from a frontal pose to an extreme lateral head pose might cause the ROI area to contain background that will suddenly shift the signal's amplitude, introducing unwanted frequencies. This case is problematic as it propagates errors in the breathing rate estimation. To overcome these kind of errors and set a performance upper bound, we also test the case when we manually fit the nostril bounding boxes around the ROI areas provided by [1]. Even though this removes the aforementioned problems, motion blur and human error during annotation might still propagate errors to the following steps. The sensitivity of methods based on ROI detection and the need for manual annotation during training, strongly justify our automatic learning approach, which does not require human supervision and is applied directly to the full input frames.

For our baseline signal extraction and breathing rate estimation we adopt a method similar to [10]. First we compute the mean pixel value in the ROI at each frame of the video. Having a signal of sufficient length (e.g. 1000 frames) we automatically remove peaks and edges that might appear due to noisy detections and movement by applying a Difference of Gaussians filter on the whole signal, then subtracting the filtered signal from the original one. We then normalize the resulting signal by subtracting its mean and dividing it by its standard deviation. After normalization we again filter it with a band-pass filter, so that only frequencies between 0.1 Hz and 0.8 Hz remain. Finally, we compute the absolute value of the Discrete Fourier Transform and obtain the maximum response in frequency, which gives the final breathing rate. Please see Table 2 for the results of our human made ROI and RetinaFace detected ROI methods.

%------------------------------------------------------------------------
\section{Our Deep-Synthetic Approach}

Next we present our novel approach to detect vital signs in thermal videos based on VSignNet - 
a deep neural architecture, which learns from pure synthetic data and without human annotation
to predict both the intensity of the vital sign (heart or respiratory rate) and the region of interest. Explained in detail in the next Sections, the synthetic data is generated elegantly and generally, 
requiring minimal domain knowledge, namely a very loose target
frequency range. Our experiments also show strong robustness to this prior knowledge,with different ranges giving similar results.

One key novelty of our approach in the context of vital sign detection
comes from learning only from synthetic data.
There is an increasing number of works in computer vision, which learn from synthetic data,
but on other tasks, such as: motion magnification[17], optical flow estimation[18], text localization in images [24], object detection [25,26], estimate depth and safe landing areas for UAVs [27].
Our synthetic data generation algorithm was inspired by approaches [10-12,14], 
which relied on the fluctuations on pixel intensities around the nostril
area during inspiration and expiration. Surprisingly, the same procedure used for breathing rate estimation, with absolutely no modification, 
was able to generalize very well to the other task of 
heart beat detection from thermal videos - by only changing the prior 
target frequency range for generating the synthetic training data.

%-------------------------------------------------------------------------
\subsection{Synthetic Data Generation}

The key motivation behind generating a
fully synthetic training data set is the little
publicly available data for research in the domain of vital sings monitoring.
As we show next, our generated training videos are very different from the target domain. However, they
are capable of capturing the quintessential elements of the targets, which makes training feasible and efficient.

\textbf{The basis of synthetic frames creation:}
The generator is designed around the initial idea of producing data sequences that imitate 
thermal images containing a nostril area, as it changes in size and location over time while the person breathes. This area is considered a region of interest
in many methods for respiratory signal detection[10-12,14].
Later the same generator proved effective in other learning tasks, such as heart rate prediction.
Thus, the generator creates an input sequence of synthetic grayscale frames, which represent a disjointed linear combination (explained next) of three layers: 1) a layer (I) containing the region(s) of interest - one or several blobs fluctuating in intensity, size and location over time according to a random target frequency within a given range, 2) a random noise layer (N) and 3) a distractor layer (D) - containing distractor blobs that behave similarly to the target ROIs, but at a frequency outside the prior target range. Each synthetic training input frame is paired with a synthetic ground truth tuple, containing:
1) the signal of interest, as a function of time 2) the average rate of the signal and 3) a sequence of binary maps
containing the interest blobs, exactly as they appear in layer I (point 1, from the synthetic input given). 

An example synthetic training input sequence can be observed in Figure 2.
Thus, the input data is a sequence of grayscale frames, containing one or several regions (objects) of interest whose
pixel intensities fluctuate between a minimum and a maximum value - interval chosen at random in [0,1],
in the same rhythm as the target signal: e.g.
object is of intensity A at the maximum of the signal, and intensity B at the 
minimum of the signal. The objects of interest are ellipses of different sizes scattered around the image, all being in sync with the same signal.

\begin{figure}[]
  \includegraphics[width=\linewidth,height=9cm]{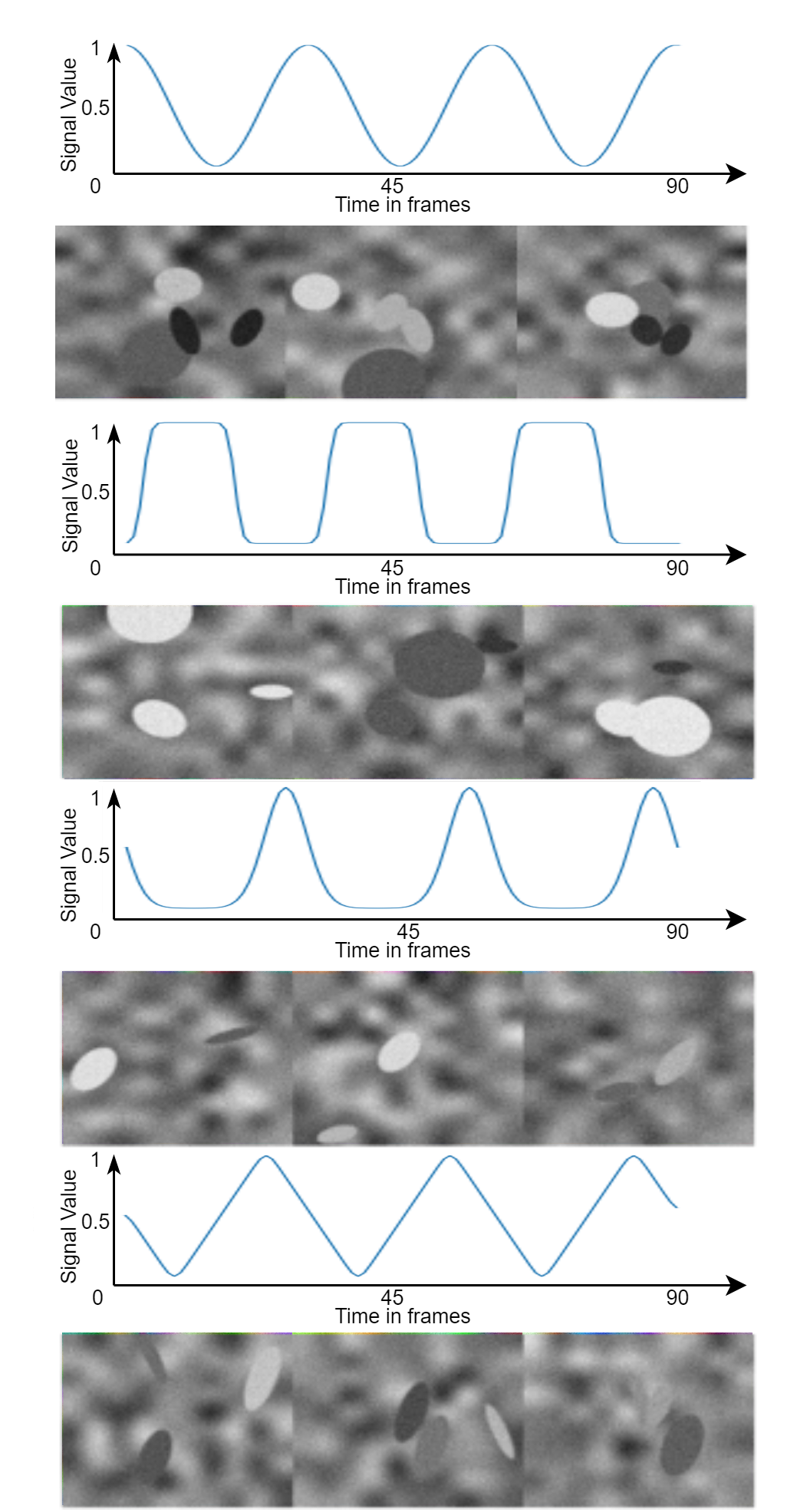}
  \captionsetup{justification=centering}
  \caption{
  Examples of generated frmes, varying according to the different types of signal of interest, which dictates the intensity of the regions of interest (I).}
  \label{fig:signals}
\end{figure}

Formally, each training videos is a sequence of T frames. For every moment $t$ a frame, $F(t)$ (Eq. 1) is a combination of signals $S_I$ and $S_D^k$, belonging to objects of interest in the set I and distractor objects in the set D. The value of the signal is multiplied with binary position masks $M_I^k$ and $M_D^k$, and then combined with a gradually shifting background, constructed based on
the background mask $M_{BG}$. The background mask is the complement of the union of all the other masks 
$M_{BG}(t)= \mathbb{C}_{\bigcup_{\substack{i \in I \cup D}} M^k(t)}$, and background intensities $\beta_{BG}$,  which are sampled from interval $\mathcal{U}[0,1]$. We first form an image of lower resolution of background intensities, which we then upsample before forming the final background mask $M_{BG}$. Then the resulting frame is first filtered with a Gaussian parameterized by $\sigma$, $G_\sigma$, before a final salt and pepper noise $\beta_{SP}$ is added.

The nature of the signal of interest $S_I$ is sampled from a selected family of cyclic functions [Sin,Step,Triangle,Gaussian], with a cycle period sampled from $\mathcal{U}[min_I,max_I]$. The distractor signals are constructed similarly, sampled from $\mathcal{U}[\mathbb{R} \setminus [min_I,max_I]]$ 

\begin{figure*}[t!]
  \includegraphics[width=\linewidth,height=7cm]{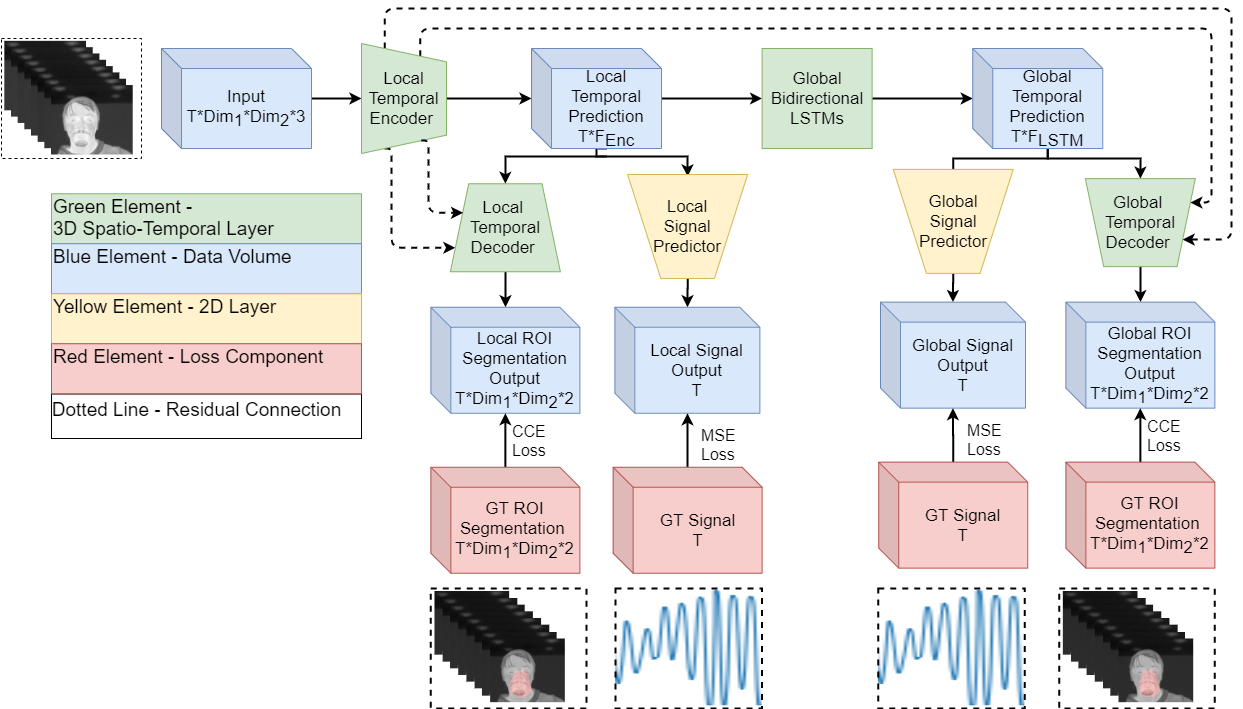}
  \caption{A high level overview of our architectural layout, displaying the general structure and shapes of
  tensors. While the specifics of each module are flexible, a key element of the architecture is the special attention 
  towards temporal information by processing at two scales, with a local and a global temporal processing pathway.}
  \label{fig:boat2}
\end{figure*}

\begin{equation}
\begin{split}
F(t)= 
&G_\sigma( \sum_{\substack{k \in I}}(S_I(t)M_I^k(t)) + \sum_{\substack{k \in D}}(S_D^k(t)M_D^k(t)) \\
&+ \beta_{BG}(t)M_{BG}(t)) + \beta_{SP}(t) \\
\end{split}
\end{equation}

\textbf{Objects and distractors moving in time and space:}
The masks are generated by drawing ellipses (originally inspired from nostrils' shapes).
The ellipses from the first frame is parameterized by a position $(p_0^k,p_1^k)$ sampled from $\mathcal{U}[(0,Dim_{Frame}^0),(0,Dim_{Frame}^1)]$, a pair of dimensions for the axes $(d_0^k,d_1^k)$ sampled from $\mathcal{U}[(1,Dim_{Frame}^0/4),(1,Dim_{Frame}^1/4))]$, and a rotation angle $\alpha$ sampled from $\mathcal{U}[0,360]$. A secondary position is sampled as well, which represents the final destination of the ellipse, $(fp_0^k,fp_1^k)$. Afterwards, at each frame, a
new size is calculated by applying slight fluctuations to the previous size $ d_i^k(t) = d_i^k(t-1) * \delta$, $\delta$ being sampled from
$\mathcal{N}(1,0.1)$. The new angle is calculated by similarly, $\alpha^k(t) = \alpha^k(t-1) * \delta$, $\delta$ being sampled from
$\mathcal{N}(1,0.1)$. The new position is calculated as the weighted average between start position and end position, over which a position noise,$\delta$, is added $p_i^k(t) = \frac{T-t}{T} p_i^k(t-1) + \delta + \frac{t}{T} \;fp_i^k$. 

\begin{equation}
\begin{split}
M^k(t)
=& Ellipse((p_1^k,p_2^k),(d_1^k,d_1^k),\alpha^k) \setminus\bigcup_{\substack{i \in I \cup D \\ o_i<o_k}}M^i(t) \\
\end{split}
\end{equation}

\textbf{Modeling the target signal:}
The target signal is composed of periodic functions (Fig. 3) with values between 0 and 1, and 
a function period sampled from the interval of interest specific to the task at hand.
This interval of interest represents the prior of the synthetic data generation method. 
It is key to selecting the signal source from among other candidates irrelevant to our task.

\textbf{Synthetic data generation summary:}
Along with the object of interest, the input data also contains the following forms of augmentation and
noise, designed to make the network robust and generalize beyond the
particular shape and appearance of the regions of interest in the ground truth:

\begin{enumerate}
\item Background level augmentations specific to the target domain: salt and pepper noise, designed to mimic the camera noise and a smoothly varying background, designed to mimic slight local changes. 
\item Object level augmentations were used: size noise, designed to simulate slight changes in scale; position noise, designed to simulate slight movements of the head and position change, designed to simulate big movements, such as head rotations. 
\item Signal level augmentations are applied as well: signal noise, designed to emulate different noise type present in the real domain and signal flattening, designed to emulate periods when the vital signs are missing. This augmentation has the effect of smoothing and cleaning the prediction of the network. 
\item An important augmentation of the data is the addition of distractor objects,
which look the same as the object of interest, but the signal frequencies are
sampled from very different frequency intervals. This is a key component when 
dealing with the presence of multiple different signals in a video.
\end{enumerate}

%-------------------------------------------------------------------------
\subsection{VSign-Net: Our Deep Learning Architecture}

We propose VSignNet, our deep learning architecture (Fig. 4), which captures the temporal dimensions
of the data on two levels, starting from a first local one and followed by a second, global one.
The data pipeline is composed of 5 types of components: Temporal Convolutional Encoder [19], Bidirectional LSTM [20], Fully Connected Predictor,Temporal Convolutional Decoder [20], Signal Analysis Module. The Fully Connected Predictor and Temporal Convolutional Decoder are both present before and after the bidirectional LSTM, capturing temporality on both a local and a global scale.

\textbf{Temporal Convolutional Encoder.} Applied on the sequence of input frames,
it encodes spatial and temporal information together, creating a powerful
embedding containing information about the slight fluctuations of the input.

Designed with a relatively small temporal receptive field, it is capable to capture local data evolution, as indicated by
the values of the auxiliary local temporal loss.  In our experiments a simple encoder was employed, consisting of 6 blocks containing Conv3D(kernel:3,stride:2,filters:64)-RELU-BatchNorm [21]-Dropout [22].

\textbf{Bidirectional LSTM.} Applied on the embedding resulting from the temporal
encoder, it has the role of aggregating global temporal information and correcting
the local information aggregated by the encoder. In our experiments two stacked bidirection LSTMs with 512 units each were used.

\textbf{Fully Connected Modules.} Present twice in the architecture, before
and after the LSTMs, having the role of transforming the embedding of each frame
in a single numerical value representing the magnitude of the vital sign at each frame. Our experiments used 3 Fully Connected layers, with 32, 8 and respectively 1 unit.

\begin{figure}[t!]
  \includegraphics[width=\linewidth]{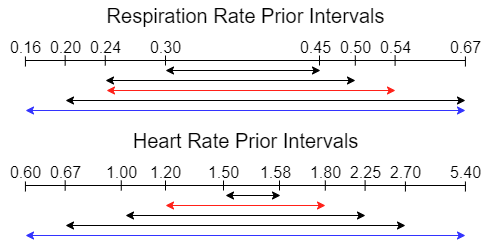}
  \caption{Selection of frequency range priors (of the signal of interest) 
  used for synthetic data generation. Red lines represent intervals [$min_{dataset}$,$max_{dataset}$], blue lines represent intervals calculated based on
  medical domain knowledge, while black lines represent intervals constructed by
  expanding and shrinking the red interval.}
  \label{fig:priors}
  \vspace{-4mm}
\end{figure}

The first is applied on the embedding resulted from the temporal encoder, containing only local temporal information. The second is applied on the embedding resulted
from the LSTMs, containing global temporal information.

\textbf{Temporal Convolutional Decoder.} Present twice in the architecture, before
and after the LSTMs, having the role of transforming the embedding of each frame
in a heatmap encoding the location of the signal source. 

The first one is applied on the embedding resulted from the temporal encoder, containing only local temporal information. The second one is applied on the embedding resulted
from the LSTMs, containing global temporal information. In our experiments a simple encoder was employed, consisting of 6 blocks containing TransposedConv3D(kernel:3,stride:2,filters:64)-RELU-BatchNorm-Dropout.

\textbf{Signal Analysis Module.} Applied on the predicted signal based on global
temporal information, it converts the signal to a numeric value representing the
frequency of the target signal.

Given the smoothness of the network's predictions, a peak detector based on local
maxima was sufficient, having the advantage of its decisions being more transparent. As in
[11], we set a minimum distance between peaks, having selected 40 frames, which is far bellow the average of an adult breathing rate. Another eligible candidate was Fourier frequency analysis[23], as applied in other methods[10,12].

%-------------------------------------------------------------------------
\section{Experimental Analysis}

Our method has been evaluated on the LCAS thermal dataset ([10]). We perform an ablation study regarding the sensitivity of the prior
on this dataset as well as experimental comparisons with our strong baseline and the methods published in the literature.
LCAS consists of 5 videos of 5 subjects, who exhibit a
regular breathing pattern. Each videos has about 2 minutes in length, captured at a 27 Hz sampling rate, with a resolution of 
382 x 288. The subjects sit still in front of the camera for approximately half the length of the video,
and for the second half they start moving closer and further, and change their head pose to extreme 
positions. The videos have ground truth annotations for breathing rate and heart rate. Breathing rate ground truth has inspiration start moments annotated, and heart rate ground truth is provided by a heart rate monitoring device. 

Evaluation has been done in the same manner as in [10], by 
counting the number of vital sign cycles in a window of time. For
respiration rate, the window is of length 1000 frames, and for heart rate
it is of 250 frames, representing about 36 and 9 seconds, respectively. 
The evaluation is split in two sections, moving and still, depending on the head movement of the subjects. 
The standard evaluation metrics, also used in LCAS[10], are Mean Absolute Error (MAE) and its Standard Deviation (STD), reported per windows of a minute or 1620 frames.
We also evaluate the temporal localization of our respiratory signal prediction, by measuring the distance between
the predicted inspiration peaks (which can be easily detected) and the human annotated inspiration start points on LCAS (Table 3 and Figure 7).

We also introduce an in-house dataset, displaying different breathing patterns, absent in LCAS[1]. The person presents four breathing patterns, Normal Nose Breathing, from frame 1 to 1740, Hold Breath, from frame 1741 to 2610,  Mouth Breathing, from frame 2611 to 3480 and Mouth and Nose Breathing, from frame 3480 to 4350. Results on this dataset are presentented in Section 4.2.

\begin{figure}[t!]
  \includegraphics[width=\linewidth]{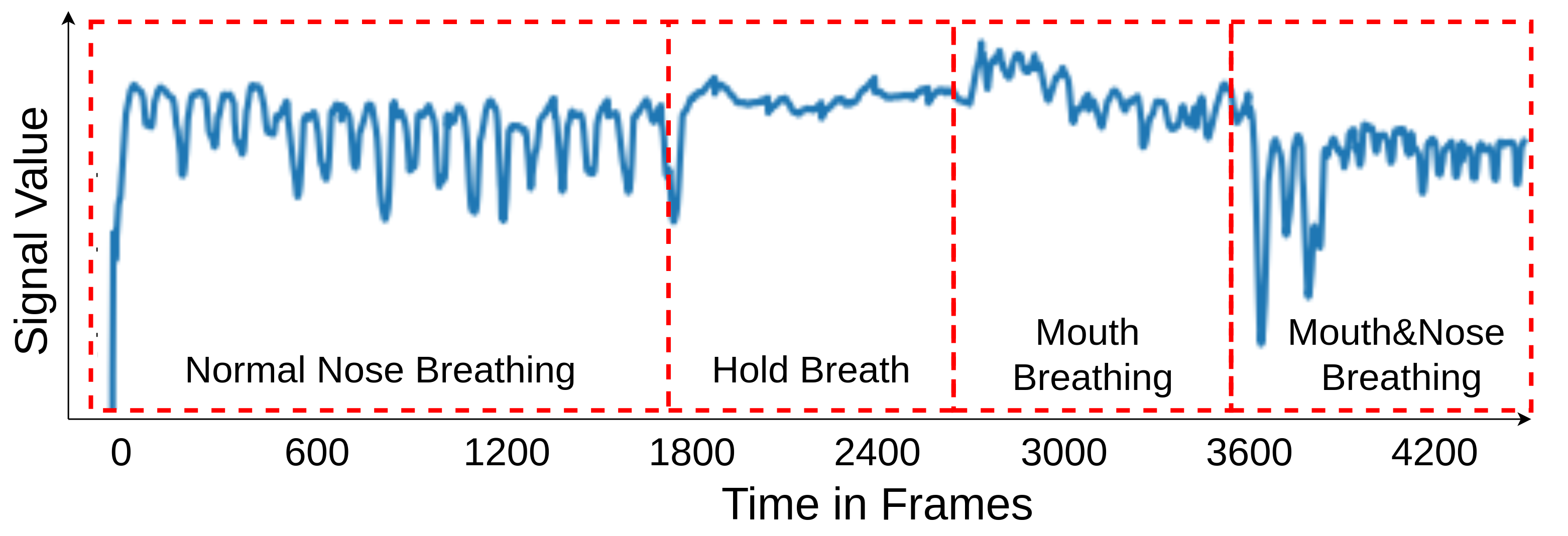}
  \caption{Example breathing signal predicted on a video featuring different breathing patterns. Changes in the predicted 
  signal patterns could be observed for different kinds of breathing (Nose breathing, Hold breath, Mouth breathing,
  Mouth and Nose breathing).}
  \label{fig:boat3}
  \vspace{-5mm}
\end{figure}

\subsection{The effect of prior target frequency range}

Considering the importance of the frequency range prior, an ablation study have been done
to test the impact of the precision of the frequency prior, knowing from annotations
the frequency intervals of interest in the LCAS[10] dataset case. At the same time
priors based on domain knowledge information, agnostic of dataset, have been tested.

As it can be observed from the results of the ablation study regarding the quality
of the prior in Table 1, obtaining a good prior can improve the performance of our method. At
the same time, less precise priors, with wider or smaller ranges than the range of the 
distribution of frequencies observed in the data, still obtain good results, indicating an ability of generalization to
proximal, but unseen frequencies. The superior results on Moving section could be attributed to the improved visibility of superficial blood
vessels [33] in lateral views.

% \subsection{VSign-Net Runtime benchmark}

% Given the practical nature of the problem, the model's runtime has been bench marked
% on a couple of platforms. As it can be observed in Table 3, the networks have a low 
% runtime with respect to the length of the input sequence. Given the need for an inference
% when the current prediction is not current anymore, making them usable in
% a real time context, even on embedded platforms such as Nvidia Jetson TX2[21].

% \vspace{0.2cm}
% \begin{table}[h!]
% \centering
% \begin{tabular}{ |p{2cm}||p{2.5cm}|p{1.5cm}|  }
%  \hline

%  Input Size & Hardware  &Runtime\\
%  \hline
%  270x112x112  &GTX1080ti    &89ms \\
%  270x112x112  &Intel i5  &870ms\\
%  270x112x112  &Jetson TX2 GPU  &1\\
%  270x112x112  &Jetson TX2 CPU  &1\\
%  \hline
% \end{tabular}
% \caption{Runtimes Benchmark done on multiple devices, including an embedded platform.}
% \label{table:2}
% \end{table}

\subsection{Predicting different breathing patterns}

In order to test the behaviour of our method when encountering the edge case of persons not breathing, we used an in-house thermal video. 
As seen in Fig. 6, the variance of the signal is much lower on 
the second region (not breathing), making it possible to detect periods 
of time with no inspirations. Also, the predictions on the third and fourth periods, mouth breathing, mouth and
nose breathing, respectively, are similar in quality to the ones predicted over the first region (nose breathing), making the
the model robust to all types of breathing.

\subsection{Experimental comparisons on LCAS dataset}
\begin{figure}[t!]
  \includegraphics[width=\linewidth,height=4cm]{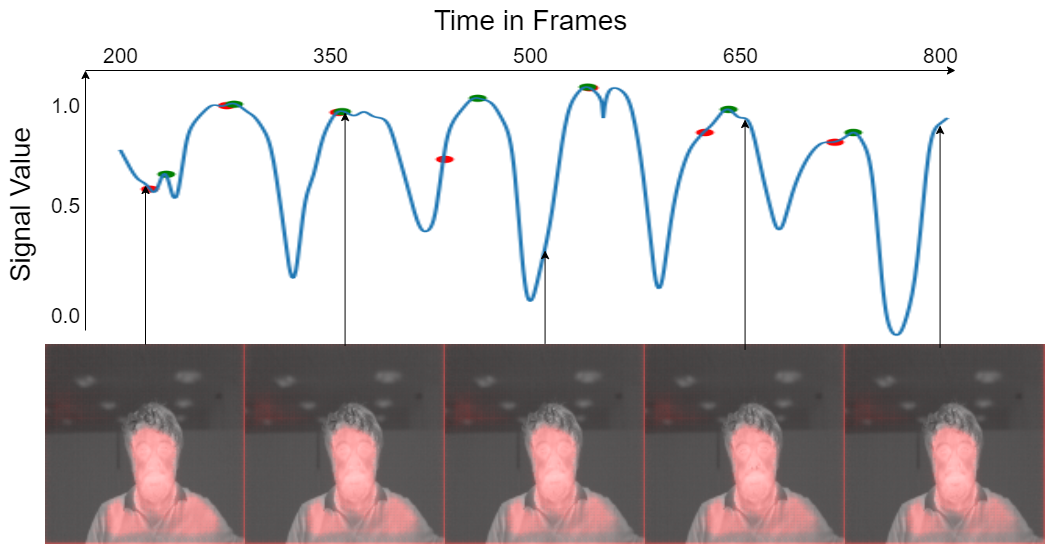}
  \caption{\textbf{Top row:} respiratory signal prediction as a function of time. In our case, the peaks of the signal (green dots) represent the moments when the inspiration reaches its maximum. The manually annotated ground truth (red dots) mark the starts of these periods of maximum inspiration, as seen by the human annotator (as expected, there is a slight misalignment between these two moments). \textbf{Bottom row:} ROI segmentation prediction in red - the region in the image likely to contain the source of the respiration signal.}
  \label{fig:RealResults}
  \vspace{-2mm}
\end{figure}

% \vspace{-0.2cm}
\begin{table}[t!]
\centering
\begin{tabular}{ |p{2.1cm}||p{0.8cm}|p{0.8cm}||p{0.8cm}|p{0.8cm}|  }

  \hline
 \multicolumn{5}{|c|}{Respiration Rate (BPM)} \\
 \hline

 \multicolumn{1}{|c||}{\multirow{2}{*}{\parbox{3cm}{\centering Trained Freq. Interval}}} & \multicolumn{2}{|c||}{\multirow{1}{*}{Still}} & \multicolumn{2}{|c|}{\multirow{1}{*}{Moving}} \\\cline{2-5}
 
  &MAE  &STD &MAE  &STD\\
 \hline
 {[0.30, 0.45] Hz}  &1.75 &$\pm$1.6 &2.24 &$\pm$1.7\\
 {[0.24, 0.50] Hz}  &2.10 &$\pm$1.7 &3.13 &$\pm$2.5\\
 \textcolor{red}{{[0.24, 0.54] Hz}}  &\textcolor{red}{1.12} &\textcolor{red}{$\pm$1.3} &\textcolor{red}{2.62} &\textcolor{red}{$\pm$2.0}\\
 {[0.20, 0.67] Hz}  &2.47 &$\pm$1.8 &3.23 &$\pm$1.9\\
 \textcolor{blue}{[0.16, 0.67] Hz}  & \textcolor{blue}{1.85} & \textcolor{blue}{$\pm$1.5} & \textcolor{blue}{3.45} & \textcolor{blue}{$\pm$2.6}\\
 \hline
  \hline
 \multicolumn{5}{|c|}{Heart Rate (BPM)} \\
 \hline
  \multicolumn{1}{|c||}{\multirow{2}{*}{\parbox{3cm}{\centering Trained Freq. Interval}}} & \multicolumn{2}{|c||}{\multirow{1}{*}{Still}} & \multicolumn{2}{|c|}{\multirow{1}{*}{Moving}} \\\cline{2-5}
  &MAE  &STD &MAE  &STD\\
 \hline
 {[1.50, 1.58] Hz}  &14.38  &$\pm$11.7 &13.98 &$\pm$9.9\\
 \textcolor{red}{[1.20, 1.80] Hz}  &\textcolor{red}{17.64}  &\textcolor{red}{$\pm$9.9} &\textcolor{red}{14.91} &\textcolor{red}{$\pm$12.1}\\
 {[1.00, 2.25] Hz}  &15.5  &$\pm$10.6 &11.18 &$\pm$7.9\\
 {[0.67, 2.70] Hz}  &17.2  &$\pm$12.7 &12.25 &$\pm$8.2\\
 \textcolor{blue}{[0.60, 5.40] Hz}  & \textcolor{blue}{15.38}  & \textcolor{blue}{$\pm$10.8} & \textcolor{blue}{14.18} & \textcolor{blue}{$\pm$11.4}\\
 \hline
\end{tabular}
\caption{Ablation study: mean absolute errors (MAE) and standard deviations (STDs), per minute, for different
signal frequency interval priors used and for different cases, Still head pose vs Moving head pose.
Red intervals are computed from dataset statistics [$min_{data}$,$max_{datas}$] and blue intervals are obtained from medical domain knowledge.}
\label{table:3}
\vspace{-4mm}
\end{table}

% \vspace{0.2cm}
\begin{table}[]
\centering
\begin{tabular}{ |p{2cm}||p{1cm}|p{1cm}||p{1cm}|p{1cm}|  }
 \hline
 \multicolumn{5}{|c|}{Heart Rate (BPM)} \\
 \hline
  \multicolumn{1}{|c||}{\multirow{2}{*}{Experiment}} & \multicolumn{2}{|c||}{\multirow{1}{*}{Still}} & \multicolumn{2}{|c|}{\multirow{1}{*}{Moving}} \\\cline{2-5}
  
  &MAE  &STD &MAE  &STD\\
 \hline
 LCAS [10]  &29.68 &$\pm$15.76 &18.96 &$\pm$22.51\\
 VSignNet  & \textbf{15.51} &$\pm$ \textbf{9.93} & \textbf{14.91} &$\pm$ \textbf{7.99}\\
 \hline
  \hline
 \multicolumn{5}{|c|}{Respiration Rate (BPM)} \\
 \hline

 \multicolumn{1}{|c||}{\multirow{2}{*}{Experiment}} & \multicolumn{2}{|c||}{\multirow{1}{*}{Still}} & \multicolumn{2}{|c|}{\multirow{1}{*}{Moving}} \\\cline{2-5}
 
  &MAE  &STD &MAE  &STD\\
 \hline
 LCAS [10]  &3.72 &$\pm$0.78 &5.87 &$\pm$2.18\\
 M ROI  &1.87 &$\pm$2.05 &4.41 &$\pm$4.41\\
 RF ROI &1.90 &$\pm$1.72 &14.77 &$\pm$7.32\\
 VSignNet  & \textbf{1.12} &$\pm$ \textbf{1.34} & \textbf{2.62} &$\pm$ \textbf{2.07}\\
 \hline
\end{tabular}
\caption{Performance comparison between our results (VSignNet) and the results reported by Cosar et al (LCAS [10]). We also report the results of
the two baselines, using either manually annotated ROI (M ROI) or ROI detected with RetinaFace (RF ROI). MAE and STD metrics are computed as in Table 1.}
\label{table:1}
\end{table}

As seen in Table 2, our method outperforms LCAS[10] and the baselines by a good margin. Note that M ROI baseline uses manually annotated regions during both testing and training, while RF ROI uses instead the automatic
RetinaFace detector both for training and testing. Our VSignNet uses no detector and takes as input raw full thermal images.

\begin{table}[]
\begin{tabular}{|c|l|l|l|l|l|}
\hline
\multicolumn{6}{|c|}{Respiratory Signal Temporal Localization} \\ \hline
\multicolumn{3}{|c||}{Still} & \multicolumn{3}{c|}{Moving} \\ \hline
\multicolumn{1}{|l|}{Mean} & STD & \multicolumn{1}{r||}{Median} & Mean & STD & Median \\ \hline
\multicolumn{1}{|r|}{0.25} & \multicolumn{1}{r|}{$\pm$0.19} & \multicolumn{1}{r||}{0.21} & \multicolumn{1}{r|}{0.27} & \multicolumn{1}{r|}{$\pm$0.24} & \multicolumn{1}{r|}{0.21} \\ \hline
\end{tabular}
\caption{Estimating the difference between the moments when the predicted inspiration period reaches its maximum and the start of the inspiration period as marked by human annotators on LCAS. The differences are estimated as the ratio between the distance in absolute number of frames (between the two moments) and the total number of frames in that specific respiration period. We report mean values, standard deviation as well as median values for the two cases of Still head pose vs. Moving head pose. Note that the mean error of 0.25 (a quarter of the total inspiration-expiration period) is in fact expected, intuitively, between the start at the peak of the inspiration period.}
\label{table:1}
\vspace{-4mm}
\end{table}

\subsection{Region of interest (ROI) segmentation}

We also introduce the prediction of the region of interest, as a ROI segmentation.  
Besides segmenting the probable source region of the signal, it could help us 
better understand the physiological cues used by the model.
In order to evaluate our ROI segmentation predictions, we use qualitative results such as the ones
displayed in Figure 7. At the same time, due to the lack of ground truth, we introduce our
own quantitative metrics, which are based on our head detections using RetinaFace. We use the following metrics: 1) Intersection over Union (IOU) with the previously mentioned detections; 
2) CHR, center Hit Rate, that is the percent of time the center of the predicted region is inside the bounding box detected with RetinaNet; 
3) DC, distance between the prediction center and the center of the face box. Evaluation of the segmentation task has been done only on the respiratory signal estimation task, where we know the main source of the signal (nostrils).

In Table 4 we show quantitative ROI segmentation results. Qualitative results at two scales are presented in Figure 8.
At the original scale, the full image is given to VSignNet, which predicts as region of interest the full face of the person. A bounding box is taken around that region and fed again into the net, which predicts, at the second, smaller scale,
regions of interest around the nose and mouth regions - proving that VSignNet has actually learned by itself to detect the primary source regions of the signal of interest.

\begin{figure}[t!]
  \includegraphics[width=\linewidth,height=3.5cm]{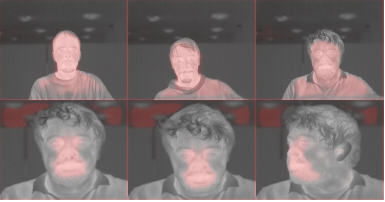}
  \caption{Signal source ROI mask shown in red, performed at two scales. At the larger scale (top), our VSignNet highlights the person, while at the smaller scale (bottom), VSignNet is capable to correctly isolate the nose and mouth regions, as primary signal sources for respiration.}
  \label{fig:RealResults}
\end{figure}

\vspace{0.2cm}
\begin{table}[]
\centering
\begin{tabular}{|p{1.6cm}||p{1.6cm}|p{1.6cm}|p{1.8cm}|}
 \hline
 \multicolumn{4}{|c|}{Respiratory ROI Segmentation Evaluation} \\
 \hline

 \multicolumn{1}{|c||}{\multirow{1}{*}{\parbox{1cm}{\centering Region}}} 
 
  &IOU ($\%$)  &CHR ($\%$) &DC (pixels)\\
 \hline
 {Head ROI} &50.95 &92.37 &6.74\\
 \hline
\end{tabular}
\caption{Quantitative evaluation of respiratory ROI segmentation 
on the LCAS dataset, obtained by VSignNet. 
The metrics are averages over LCAS: intersection over union (IOU), center hit rate (CHR) and distance between the detection and ground truth centers of mass (DC).}
\label{table:3}
\vspace{-4mm}
\end{table}

%-------------------------------------------------------------------------
\section{Conclusions}

We presented a novel deep learning approach with fully automated synthetic training
for detecting vital signs and their source interest regions in thermal videos. Different from the published literature
our method employs a novel deep neural net (VSignNet), with two, local and global, temporal stages of processing, which 
achieves state-of-the-art results on the recent LCAS dataset. 
Our second contribution is that we overcome the lack of proper supervised training data with an elegant and general algorithm for synthetic training data generation. Our method is based on minimum prior medical knowledge and it is applicable (without modification) to both heart and respiratory rate estimation, as our experiments show. It is truly interesting that a very general and relatively simple algorithm for generating synthetic training data can be successfully applied in the complex and specific domain of medical imaging. This fact opens up new questions, with broader impact, regarding the ability of such strategy to learn, without human supervision, other complex vision tasks in space and time.

\textbf{Acknowledgements}. We thank Advanced Camera Laboratory, LG Electronics, Seoul, Korea, for their support and collaboration.

%-------------------------------------------------------------------------
\section{References}

\begin{enumerate}[{label=[\arabic{*}]}]

\item L Tarassenko et al, “Non-contact video-based vital sign monitoring using ambient light and auto-regressive models” , Physiol. Meas. 35 807, (2014)
\item  Mayank Kumar, Ashok Veeraraghavan, and Ashutosh Sabharwal, "DistancePPG: Robust non-contact vital signs monitoring using a camera," Biomed. Opt. Express 6, 1565-1588 (2015)
\item  A. Procházka et al, “Breathing Analysis Using Thermal and Depth Imaging Camera Video Records” Sensors, 17, 1408 (2017)
\item  F. Q. AL-Khalidi, R. Saatchi, D. Burke, and H. Elphick. Tracking
human face features in thermal images for respiration monitoring. In
AICCSA 2010, pages 1–6, May 2010.
\item  C. B. Pereira et al, “Monitoring of Cardiorespiratory Signals Using Thermal Imaging” Sensors, 18, 1541 (2018)
\item  Youngjun Cho, et al,  “Instant Automated Inference of Perceived Mental Stress through Smartphone PPG and Thermal Imaging” Journal of Medical Internet Research (JMIR) Mental Health - Special Issue on Computing and Mental Health (2018)
\item  Paul E.H. Ricard, in Acute Care Handbook for Physical Therapists (Fourth Edition), 2014
\item  C. B. Pereira, X. Yu, V. Blazek and S. Leonhardt, "Robust remote monitoring of breathing function by using infrared thermography," 2015 37th Annual International Conference of the IEEE Engineering in Medicine and Biology Society (EMBC), Milan, 2015, pp. 4250-4253.
\item  A. B. Hertzman, ”Photoelectric plethysmography of the fingers and toes in man”, Exp. Biol. Med. 37(3), pp. 529-534 (1937). 

\item  S. Cosar, Z. Yan, F. Zhao, T. Lambrou, S. Yue and N. Bellotto. "Thermal camera based physiological monitoring with an assistive robot. In: \emph{IEEE International Engineering in Medicine and Biology Conference}, 17-21 July 2018
\item  K. Mutlu, J.Esquivelzeta Rabell, P. Martin del Olmo and S. Haesler. "IR thermography-based monitoring of respiration phase without image segmentation". In: \emph{Journal of Neuroscience Methods}, 301:1-8, 2018
\item  Y. Cho, S. J. Julier, N. Marquardt and N. Bianchi-Berhouze. Robust tracking of respiratory rate in high-dynamic range scenes using mobile thermal imaging". In: \emph{Biomed Opt Express}, 8(10):4480-4503, 2017
\item  Z. Kalal, K. Mikolajczyk and J.Matas. "An Iterative Image Registration Technique with an Application to Stereo Vision". In: \emph{2010 in 20th International Conference on Pattern Recognition (ICPR)}, 2010, pp. 2756-2759
\item  C. B. Pereira, X. Yu, M. Czaplik, R. Rossaint, V. Blazek and S. Leonhardt. "Remote monitoring of breathing dynamics using infrared thermography". In \emph{Biomed Opt Express}, 6(11):4378-4394, 2015

\item Deng, J., Guo, J., Zhou, Y., Yu, J., Kotsia, I., Zafeiriou, S. (2019). Retinaface: Single-stage dense face localisation in the wild. arXiv preprint arXiv:1905.00641.

\item Yang, Shuo, et al. "Wider face: A face detection benchmark." Proceedings of the IEEE conference on computer vision and pattern recognition. 2016. 
\item  T. Oh, R. Jaroensri, C. Kim, M. Elgharib, F. Durand, W. Freeman, W. Matusik "Learning-based Video Motion Magnification" arXiv preprint arXiv:1804.02684 (2018).
\item  Mayer, N., Ilg, E., Fischer, P. et al. What Makes Good Synthetic Training Data for Learning Disparity and Optical Flow Estimation?. Int J Comput Vis 126, 942–960 (2018). https://doi.org/10.1007/s11263-018-1082-6
\item Lea, Colin, et al. "Temporal convolutional networks for action segmentation and detection." proceedings of the IEEE Conference on Computer Vision and Pattern Recognition. 2017.
\item  Sepp Hochreiter and Jürgen Schmidhuber. 1997. Long Short-Term Memory. Neural Comput. 9, 8 (November 1997), 1735–1780. DOI:https://doi.org/10.1162/neco.1997.9.8.1735
\item Ioffe, Sergey, and Christian Szegedy. "Batch normalization: Accelerating deep network training by reducing internal covariate shift." arXiv preprint arXiv:1502.03167 (2015).
\item Nitish Srivastava, Geoffrey Hinton, Alex Krizhevsky, Ilya Sutskever, and Ruslan Salakhutdinov. 2014. Dropout: a simple way to prevent neural networks from overfitting. J. Mach. Learn. Res. 15, 1 (January 2014), 1929–1958.
\item Chaparro, Luis. (2015). Fourier Analysis of Discrete-Time Signals and Systems. 10.1016/B978-0-12-394812-0.00011-5. 
\item Gupta, A., Vedaldi, A. and Zisserman, A., Synthetic data for text localisation in natural images. In Proceedings of the IEEE Conference on Computer Vision and Pattern Recognition (CVPR). 2016
\item Peng, X., Sun, B., Ali, K. and Saenko, K. Learning deep object detectors from 3d models. In Proceedings of the IEEE International Conference on Computer Vision (ICCV), 2015. 
\item Tremblay, J., Prakash, A., Acuna, D., Brophy, M., Jampani, V., Anil, C., To, T., Cameracci, E., Boochoon, S. and Birchfield, S., 2018. Training deep networks with synthetic data: Bridging the reality gap by domain randomization. In CVPR Workshops, 2018.
\item Marcu, A. et al.  SafeUAV: Learning to estimate depth and safe landing areas for UAVs from synthetic data. European Conference on Computer Vision (ECCV) UAVision Workshop, 2018.
\item Takano, C., \& Ohta, Y. (2007). Heart rate measurement based on a time-lapse image. Medical Engineering \& Physics, 29(8), 853-857. https://doi.org/10.1016/j.medengphy.2006.09.006 
\item Verkruysse, W., Svaasand, L. O., \& Nelson, J. S. (2008).  Remote plethysmographic imaging using ambient light. Optics Express, 16(26), 21434. https://doi.org/10.1364/oe.16.021434 
\item Poh, M.-Z., McDuff, D. J., \& Picard, R. W. (2011). Advancements in Noncontact, Multiparameter Physiological Measurements Using a Webcam. IEEE Transactions on Biomedical Engineering, 58(1), 7–11. https://doi.org/10.1109/tbme.2010.2086456 
\item Wang, W., den Brinker, A. C., Stuijk, S., \& de Haan, G. (2017, May). Color-Distortion Filtering for Remote Photoplethysmography. 2017 12th IEEE International Conference on Automatic Face \& Gesture Recognition (FG 2017). 2017 12th IEEE International Conference on Automatic Face \& Gesture Recognition (FG 2017). https://doi.org/10.1109/fg.2017.18 
\item Chen, W., \& McDuff, D. (2018). DeepPhys: Video-Based Physiological Measurement Using Convolutional Attention Networks. In Computer Vision – ECCV 2018 (pp. 356–373). Springer International Publishing. https://doi.org/10.1007/978-3-030-01216-8\_22 
\item Garbey, M., Sun, N., Merla, A., \& Pavlidis, I. (2007). Contact-Free Measurement of Cardiac Pulse Based on the Analysis of Thermal Imagery. IEEE Transactions on Biomedical Engineering, 54(8), 1418–1426. doi:10.1109/tbme.2007.891930 
\end{enumerate}

\end{document}